# EXTRACTING USEFUL RULES THROUGH IMPROVED DECISION TREE INDUCTION USING INFORMATION ENTROPY

Mohd. Mahmood Ali[1], Mohd. S. Qaseem[2], Lakshmi Rajamani[3], A. Govardhan[4]

[1]Dept. of C.S.E., Muffakhamjah College of Engg. & Tech., Hyderabad, India
mahmoodedu@gmail.com
[2]Dept. of C.S.E., Nizam Institute of Engg. & Tech., Hyderabad, India
ms_qaseem@yahoo.com
[3]Dept. of C.S.E., Univ. College of Engg., O.U., Hyderabad, India
drlakshmiraja@gmail.com
[4]Dept. of C.S.E., JNTUH., Hyderabad, India
dejntuh@jntuh.ac.in

## ABSTRACT

*Classification is widely used technique in the data mining domain, where scalability and efficiency are the immediate problems in classification algorithms for large databases. We suggest improvements to the existing C4.5 decision tree algorithm. In this paper attribute oriented induction (AOI) and relevance analysis are incorporated with concept hierarchy's knowledge and HeightBalancePriority algorithm for construction of decision tree along with Multi level mining. The assignment of priorities to attributes is done by evaluating information entropy, at different levels of abstraction for building decision tree using HeightBalancePriority algorithm. Modified DMQL queries are used to understand and explore the shortcomings of the decision trees generated by C4.5 classifier for education dataset and the results are compared with the proposed approach.*

## 1. INTRODUCTION

Classification is a supervised learning technique in data mining where training data is given to classifier that builds classification rules. Later if test data, is given to classifier, it will predict the values for unknown classes.

C4.5 classifier [1],[2], a well-liked tree based classifier, is used to generate decision tree from a set of training examples. Nowadays C4.5 is renamed as J48 classifier in WEKA tool, an open source data mining tool. The heuristic function used in this classifier is based on the concept of information entropy.

Induction of decision trees from very large training sets has been previously addressed by SLIQ [3] and SPRINT [4], but the data stored is without generalization. The generalization concept for evaluating classification rules using DMQL in data cube is proposed [5]. Decision tree construction process performed on very large datasets leads to bushy or meaningless results. This issue which is addressed in data generalization and summarization based characterization [5], consist of three steps attribute-oriented induction, where the low-level data is generalized to high-level data using the concept hierarchies, relevance analysis [6], and multi-level mining, where decision trees can be induced at different levels of abstraction. The integration of these steps leads to efficient, high quality and the elegant handling of continuous and noisy data. An inherent weakness of C4.5 classifier, is that the information gain attribute selection criterion favors many-valued attributes, due to which some of the attributes are pruned because of less information gain in decision tree construction. Ultimately this leads to less classification rules. C4.5 classifier encounters the over-branching problem caused by unnecessary partitioning of the data which is of least importance to users. Therefore, we propose a HeightBalancePriority algorithm which constitutes NodeMerge and HeightBalance algorithms [11] that allows merging of nodes in the decision tree thereby, discouraging over-partitioning of the data. This algorithm also uses the concept of Height Balancing in the decision tree based on priority checks at every node based on attribute values selected for different attributes using information entropy. This enhances the overall performance, as the final decision tree constructed is efficient enough to derive the complete classification rules effectively avoiding over-branching problem of unnecessary attributes.

In this paper, we emphasized on decision tree construction and classification rules derived based on priority by using information entropy, which gives importance to selected attributes that was neglected by C4.5 classifier. In the Remainder of the paper, for clarity and better understanding, the Modified DMQL queries are used that explore the dataset used for decision tree construction.

The paper is organized as follows. Section 2, describes the data generalization and summarization based characterization. Section 3 briefly, explains about the proposed algorithms used for decision tree construction. The Decision tree formed using C4.5, yet suffers from deficiencies. To overcome the priority based decision tree construction method is discussed in Section 4. In Section 5, some more examples are taken to explore C4.5, deficiencies and results obtained are compared with proposed approach. Section 6 summarizes the methodology for priority based Decision tree construction and future challenges.

## 2. CLASSIFICATION USING DECISION TREE INDUCTION

Scalability and efficiency issues for classification techniques are addressed for large databases that has improved C4.5 classifier, using the following four steps:
- Generalization by AOI, which compresses training data. This includes storage of generalized data in data cube to allow fast accessing [8].
- Relevance analysis, that removes irrelevant attributes, thereby, further compacting training data.
- Multi-level mining, which combines the induction of decision trees with knowledge in concept hierarchies.
- Priority based height balance trees using Entropy.

The top three methodologies are discussed in Generalization and Decision tree induction for efficient classification [7], we have also used the above three methodologies with slight improvements, where as the priority based decision tree construction using Information entropy, is discussed in Section 4.

### 2.1. Generalization by AOI (Attribute oriented Induction)

AOI [9], a knowledge discovery tool which allows the generalization of data, offers two major advantages for the mining of large databases. Firstly, it allows the raw data to be handled at higher

International Journal of Information Sciences and Techniques (IJIST) Vol.3, No.1, January 2013conceptual levels. Generalization is performed with the use of attribute concept hierarchies, where the leaves of a given attribute's concept hierarchy correspond to the attribute's values in the data (referred to as primitive level data) [10]. Secondly, generalization of the training data is achieved by replacing primitive level data by higher level concepts. Hence, attribute- oriented induction allows the user to view the data at more meaningful abstractions.

Furthermore, AOI, addresses the scalability issue by compressing the training data. The generalized training data will be much more compact than the original training set, and hence, will involve fewer input/output operations. With the help of AOI, many-valued attributes in the selection of determinant attributes are avoided since AOI reduces large number of attribute values to small set of distinct values according to the specified thresholds.

AOI also performs generalization by attribute removal [9]. In this technique, an attribute having a large number of distinct values is removed if there is no higher level concept for it. Attribute removal further compact the training data and reduces the bushiness of resulting trees. Hence, aside from increasing efficiency, AOI may result in classification trees that are more understandable, smaller, and therefore easier to interpret than trees obtained from methods operating on non-generalized (larger) sets of low-level data. The degree of generalization is controlled by an empirically set generalization threshold. If the number of distinct values of an attribute is less than or equal to this threshold, then further generalization of the attribute is halted. We consider a simple example to explain all the detail steps to generalize the final classification tree and find out the classification rules. Table 1, depicts a raw training data of class of average education level is used, *region* wise, with the count in relation with the country part to which that place belongs around the world.

Table 1. Attributes with four attribute values.

| Average Education level | Region | Family Income per year | Income Level |
|---|---|---|---|
| Illiterate | Cuba.north | $ 899 | Low |
| fouryearscollege | USA.east | $ 30000 | Medium |
| Fouryearscollege | USA.south | $ 38000 | High |
| Fouryearscollege | USA.middle | $ 32000 | High |
| twoyearscollege | USA.middle | $ 30400 | High |
| Graduate school | China.south | $ 38000 | High |
| Elementary school | Cuba.north | $ 990 | Low |
| High school | India.east | $ 7839 | Low |
| Fouryearscollege | China.east | $ 30000 | Medium |
| Graduate school | China.west | $ 38000 | High |
| Junior High | China.south | $ 3800 | Low |
| Twoyearscollege | India.south | $ 20000 | Medium |
| Fouryearscollege | USA.west | $ 20000 | Medium |
| Graduate school | China.west | $ 38000 | High |
| Ph. D | India.south | $ 50000 | High |

The Generalization using AOI for the attribute *WORLD* is depicted in Figure 1. Similarly the concept hierarchies for age, income, and education can also be represented. We illustrate ideas of attribute-oriented induction (AOI) with an example for edu_dataset shown in table 1, generalized using the concept hierarchy. AOI is performed on set of relevant data. An intermediate relation, achieved by concept ascension using concept hierarchies as shown in Figure 1, attribute *region* normalized to country as shown in Table 2. Owing to attribute removal technique, attribute values INDIA.EAST, INDIA.WEST, INDIA.SOUTH, are replaced with INDIA, as depicted in Figure 1.

{CUBA,CHINA,USA,INDIA} ⊂ {WORLD};
{CUBA.NORTH, CUBA.SOUTH} ⊂ {CUBA};
{USA.EAST,USA.WEST,USA.SOUTH,USA.MIDDLE,} ⊂ {USA};
{CHINA.SOUTH,CHINA.EAST,CHINA.WEST} ⊂ {CHINA};
{INDIA.EAST,INDIA.SOUTH,INDIA.WEST} ⊂ { INDIA};

Figure 1. Concept hierarchy for attribute values of *region*

Identical tuples for Table 1 merged while collecting the count information shown in Table 2, which is achieved by applying modified data mining query language (DMQL) queries as shown with example 2.1. The resultant table obtained is substantially smaller than the original table, generalized with task-relevant data as shown in Table 2.

Example 2.1. Classification task.

**Classify** Region_Data **till** World_Data **replace** {USA,Cuba,India,China} **attribute_values with new_attribute** *Country*
**in relevance to** *avg_edu_level,country* **new_attribute** *count*
**from** edu_dataset
**where** Country_Data={"India","USA","Cuba","China"}
**and** Region_Data={"India.east","India.west","India.south",
"USA.east","USA.west","USA.south"," USA.middle","Cuba.north",
"Cuba.south", "China.south","China.east","China.west"}

In the above modified DMQL query the bold words represent reserve keywords. The keyword Classify constitutes the proposed approach algorithms, which are discussed in this paper.

Table 2. Generalized data set obtained after AOI using DMQL query of example 2.1 with count as new attribute.

| Average Education level | Country | Family Income per year | Count |
|---|---|---|---|
| Illiterate | Cuba | $1899 | 2 |
| FourYearsCollege | USA | $120,000 | 4 |
| GraduateSchool | China | $114,000 | 3 |
| TwoYearsCollege | India | $40,000 | 2 |
| FourYearsCollege | China | $30000 | 1 |

## 2.2. Relevance Analysis

The uncertainty coefficient U(A) for attribute A is used to further reduce the size of the generalized training data as shown in Equation (1). U(A) is obtained by normalizing the information gain of A so that U(A) ranges from 0 (meaning statistical independence between A and the classifying attribute) to 1 (strongest degree of relevance between the two attributes). The user has the option of retaining either the n most relevant attributes or all attributes whose uncertainty coefficient value is greater than a pre-specified uncertainty threshold, where n and the threshold are user-defined. Note that it is much more efficient to apply the relevance analysis to the generalized data rather than to the original training data.

$$U(A) = \frac{I(p_1, p_2, \ldots, p_m) - E(A)}{I(p_1, p_2, \ldots, p_m)} \qquad \text{Equation\ldots (1)}$$

Where: $I(p_1, \ldots, p_m) = -\sum^m p_i/p \, \log_2 p_i/p$

$$E(A) = \sum_{j=1}^{p} \frac{p_{1j} + \ldots + p_{mj}}{p} I(p_{1j}, \ldots, p_{mj}) \qquad \text{Equation\ldots (2)}$$

$$Gain(A) = I(p_1, p_2, \ldots, p_m) - E(A) \qquad \text{Equation\ldots (3)}$$

Here P is the set of the final generalized training data, where P contains m distinct values defining with the output distinct output class $P_i$ (for i = 1, 2, 3,…,m) and P contains pi samples for each Pi, then the expected information i.e. *Information Entropy* needed to classify a given sample is $I(p1, p2,\ldots, p_m)$. For example: we have the attribute A with the generalized final value $\{a_1, a_2, a_3, \ldots, a_k\}$ can be partition P into $\{C_1, C_2, C_3, \ldots, C_k\}$, where $C_j$ contain those samples in C that have value $a_j$ of A. The expected information based on partitioning by A is given by E(A) Equation (2), which is the average of the expected information. The Gain(A) given in Equation (3), is the difference of the two calculations. If the uncertainty coefficient for attribute A is 0, which means no matter how we partition the attribute A, we can get nothing lose information. So the attributes A have no effect on the building of the final decision tree. If U(A) is 1, mean that we can use this attribute to classify the final decision tree. This is similar to find the max goodness in the class to find which attribute we can use to classify the final decision tree. After the relevance analysis, we can get rid of some attribute and further compact the training data based on selection of attributes by users.

Entropy for *income* class which has attribute values *(High, Medium, Low)*:

$I(s_1, s_2, s_3) = I(5,5,7) = -(\frac{5}{17}) \log_2 \frac{5}{17} - (\frac{5}{17}) \log_2 \frac{5}{17} - (\frac{7}{17}) \log_2 \frac{7}{17}$

$= [-0.29411764705882 * \log_2(0.29411764705882) -$
$\quad 0.29411764705882 * \log_2(0.29411764705882) -$
$\quad 0.41176470588235 * \log_2(0.41176470588235)]$

$= 1.56565311164580141944$

If we want to choose one of the attribute as root node then, we have to calculate Information Entropy for all other attributes present, in the *education dataset*, in most of cases the continuous values are given preference to be the root node, if not multi-valued attribute values of an attribute is selected as root node. This depends effectively on the highest information gain for the attribute values present in the attributes (*Average education level, Region, family income per year*). Let us calculate the information gain for all other attributes using equation (1) and equation (2) discussed above.

We have to still compute the expected information for each of these distributions

For *Avg education level = "Illiterate"*

$s_{11} = 2 ; s_{21} = 0; \& s_{31} = 0$

$I(s_{11}, s_{21}, s_{31}) = -(\frac{2}{2}) \log_2 \frac{2}{2} = 0$

For *Avg education level = "4yearscollege"*

$s_{12} = 2; s_{22} = 3; \& s_{32} = 0$

$I(s_{12}, s_{22}, s_{32}) = -(\frac{2}{5}) \log_2 \frac{2}{5} - (\frac{3}{5}) \log_2 \frac{3}{5}$

$\qquad = 0.97095059445469$

For *Avg education level = "2yearscollege"*

$s_{13} = 1; s_{23} = 2; \& s_{33} = 0$

$I(s_{13}, s_{23}, s_{33}) = -(\frac{1}{3}) \log_2 \frac{1}{3} - (\frac{2}{3}) \log_2 \frac{2}{3}$

$\qquad = 0.91829583405451$

For *Avg education level ="Graduate school"*
  $s_{14}=3; s_{24}=0; \& s_{33}=0$
  $I(s_{14}, s_{24}, s_{34}) = -\left(\frac{3}{3}\right)\log_2 \frac{3}{3} = 0$

For *Avg education level ="Elementary school"*
  $s_{15}=1; s_{25}=0; \& s_{35}=0$
  $I(s_{15}, s_{25}, s_{35}) = -\left(\frac{1}{1}\right)\log_2 \frac{1}{1} = 0$

Similarly entropy for attribute values of attribute *Avg education level =”Ph.D”*, "High School", "Junior High" is 0; Using the equation (2) given above, if the samples are partitioned according to *Avg education level*, then

$E(Avg\ education\ level) =$
$= \frac{2}{17} I(s11, s21, s31) + \frac{5}{17} I(s12, s22, s32) + \frac{3}{17} I(s13, s23, s33) +$
$\frac{3}{17} I(s14, s24s34) + \frac{1}{17} I(s15, s25, s35) + \frac{1}{17} I(s16, s26, s36) +$
$\frac{1}{17} I(s17, s27, s37) + \frac{1}{17} I(s18, s28, s38)$

$= 0 + 0 + \left(\frac{5}{17} * 0.970950\right) + \left(\frac{3}{17} * 0.91829\right) + 0 + 0 + 0 + 0$
$= 0.44762470588235$

Hence, the gain in information from such a partitioning would be calculated using equation (3) given above:

*Gain (Avg education level)* $=$ $I(s_1, s_2, s_3) - E(Avg\ education\ level)$
  $= 1.56565311164580141944 - 0.44762470588235$
  $= 1.11802840576345141944$

Similarly, we have calculated Gain(*Region*)=1.3; Gain(family income per year)=0; We can say that Gain is high for *Region*, so we have to select *Region* as root node. After the decision tree is built, tree pruning phase is applied, resulting in removing of the nodes (attributes) which are of user interest from the tree. Cross-validation technique analyzes the final constructed tree that estimate errors, and rebuilds the decision tree as used in WEKA tool discussed in Section 4 and Section 5. Some of the classification rules are skipped resulting in inadequate decision trees. So to avoid the above problems, in our approach, we propose algorithms discussed below (algorithm 4) which give priority preference to the nodes (attributes) for decision tree construction with hidden classification rules by swapping the priorities of attributes discussed in Section 3, and with an example in Section 4. The DMQL queries are used in this paper to understand the propose work clearly, these DMQL queries include our proposed algorithms integrated.

### 2.3. Multi-level mining

The third step of decision tree construction method is multilevel mining. This combines decision tree induction of the generalized data obtained in Sections 2.1 and 2.2 (Attribute- oriented induction and relevance analysis) with knowledge in the concept hierarchies. The induction of decision trees is done at different levels of abstraction by employing the knowledge stored in the concept hierarchies. Furthermore, once a decision tree has been derived [5], the concept hierarchies can be used to generalize individual nodes in the tree, allowing attribute roll-up or drill- down, and reclassification of the data for the newly specified abstraction level. This depends on information entropy and the selection of attributes by the user.

The main idea is to construct a decision tree based on these proposed steps and prune it accordingly based on priority using Height Balancing tree concept [14], without losing any of the classification rules. The Decision Tree Construction Algorithm 1 is shown in Figure 2, which constructs a decision tree for the given training data. The stopping criteria, depends not only on maximum information entropy but also selection of the attributes by the users and its priority. In some cases if the attributes selected by the user are not in the data set then thresholds plays a vital role as previously addressed [7] for decision tree construction. Apart from *generalization threshold*, we used two other thresholds for improving the efficiency namely, *exception threshold (∈)* and *classification threshold (κ)*. Because of the recursive partitioning, some resulting data subsets may become so small that partitioning them further would have no statistically significant basis. These "insignificant" data subsets are statistically determined by the *exception threshold*. If the portion of samples in a given subset is less than the threshold, further partitioning of the subset is halted. Instead, a leaf node is created which stores the subset and class distribution of the subset samples.

The splitting-criterion in the Algorithm 1, deals with both the threshold constraints and also information gain calculation for the data. In this process, the candidate with maximum information gain is selected as "test" attribute and is partitioned. The condition, if the frequency of the majority class in a given subset is greater than the classification threshold, or if the percentage of training objects represented by the subset is less than the exception threshold, is used to terminate classification otherwise further classification will be performed recursively until all the attributes are not selected which is of users interest. The algorithm operates on training data that has been generalized to an intermediate level by AOI, and for which unimportant attributes have been removed by relevance analysis, if attributes selected are not present in the users list.

**Algorithm 1.** Decision Tree Construction

**Input:** Pre-processed Dataset without noise.
**Output**: Final well-balanced Decision tree built based on priority attributes selected by users interest.

*DecisionTree (Node n, DataPartition D)*
*{ Apply AOI-Method to D to find splitting-criterion of node n*
   *Let k be the number of children of n if k>0 do*
      *Create k children $c_1, c_2,..., c_k$ of n*
      *Use splitting-criterion to partition D into $D_1, D_2..., D_k$*
      *for i = 1 to k do*
         *DecisionTree($c_i$, Di)*
      *end for*
   *endif*
      *//Assign priority to the nodes based on the attribute*
*Call HeightBalancePriority(HeightBalance, Priority_Attribute)*
    *//Calls Algorithm 4*
    *ALLOCATE highest information entropy attributes to PRIORITY attribute/*

*}*

Figure 2. Decision tree construction algorithm with priority on attribute values for distinct attributes selected by calling the algorithm shown in Figure 4.b

In this way, the tree is first fully grown based on these conditions. Then under the pruning process, we used two algorithms namely, *NodeMerge* and *HeightBalance* algorithms [11], which help in enhancing the efficiency in case of dynamic pruning by avoiding the bushy growth of decision trees, with *HeightBalancePriority* algorithm, gives priority and builds decision tree considering all the classification rules, that C4.5 classifier had neglected.

### 2.4. Priority based Decision tree construction

The C 4.5 classifier gives preference to multi valued attributes, for which information entropy is calculated and the attribute that has the highest information entropy is selected as root node. This continues till the decision tree is fully grown.

Similarly in this priority based decision tree construction approach, Information entropy is calculated for all attributes using the concepts explained in sections 2.1, 2.2, and 2.3. But attributes that are selected by the user is given higher preference, next the priority is considered. During this, if the attribute selected by the user has less information entropy in such a case the information entropy is to be replaced by attribute that has highest Information entropy. This continues till all the attributes (which is of users interest) are selected and replaced based on higher information entropy of other attributes which is of least important to users.

Finally the decision tree is grown with all attributes that is of users interest, with all the classification rules.

## 3. PROPOSED DECISION TREE ALGORITHMS

The Decision tree construction algorithm integrates attribute-oriented induction and relevance analysis with modified version, of the C4.5 classifier is outlined in Algorithm 1. Algorithm 2, is used for merging of nodes as decision tree is built considering all the nodes, is shown in Figure 3.

**Algorithm 2.** Merging of nodes at multiple levels of abstraction when decision tree is constructed.

**Input:** All Nodes with attributes and attribute values given
**Output**: Merged decision tree is constructed.

```
NodeMerge( NodeData_A, NodeData_B)
{ Check priorities for node_A and node_B;
            if both the priorities > checkpoint then
  { link_AB = remove_link_joining(NodeData_A, NodeData_B);
    union = NodeData_A.merge_with(NodeData_B);
    for (related_node: nodes_incident_to_either (NodeData_A, NodeData_B))
         link_RA = link_joining (related_node, NodeData_A); link_RB = link_joining
            (related_node, NodeData_B); disjoin (related_node, NodeData_A);
                      disjoin (related_node, NodeData_B);
                      join (related_node, union, merged_link);
    }
    else print (" Node have high priority, cannot be merged");
    Call HeightBalance (union, new_link_AB);  // calls Algorithm 3
}
```

Figure 3. Algorithm to merge nodes at multiple levels of abstraction during Decision tree construction.

During the process of merging the nodes, the algorithm checks for assigned priorities for attribute values of distinct attributes during decision tree construction. The primary disadvantage of ordinary tree is that they attain very large heights in ordinary situation when large data sets are given to classifier as seen in C4.5. To overcome this problem *Heightbalance* algorithm 3 is used outlined in Figure 4 (a).

Decision tree constructed is prioritized, by giving preference to specific attribute considering highest Information gain with the given algorithm 4, shown in Figure 4 (b).

**Algorithm 3.** Height balancing of decision tree.

*HeightBalance (union, link_AB)*
  *{ Check whether the tree is imbalanced or not;*
    *if yes then*
    *{ if balance_factor ( R ) is heavy*
       *{ if tree's right subtree is left heavy then*
              *perform double left rotation;*
         *else*
              *perform single left rotation;*
    *}*
        *else if balance_factor( L ) is heavy*
        *{     if tree's left subtree is right heavy then*
                 *perform double right rotation;*
              *else*
    *perform single right rotation;*
           *}*
     *}*
     *Print("Tree is balanced");*
     *Check for path preservations;*
*}*

Figure 4 (a). Height balancing for Decision tree

**Algorithm 4.** Height balancing for decision tree with priority.
**Input:**   Nodes with Height balanced tree are given
**Output**: Prioritized Height balanced decision tree.

*HeightBalancePriority (HeightBalance, Priority_Attribute)*
*{*
*Check for Priority in HeightBalance(Union,link_AB)*
    *Evaluate (priority_attribute (Attributevalue1, .., Attributevaluen))*
  *{*
       *Switch(priority_attribute)*
  *{*
   *Case 1:*
     *Priority_Attribute='Attributevalue1' //place it first*
     *Check InformationGain //calculate Information Gain*
     *Allocate the first Highest InformationGain //first priority Attrib.*
    *Case 2:*
     *Priority_Attribute='Attribute value2'  //place it second*
     *Check InformationGain //calculate Information Gain*
     *Allocate the second Highest Information Gain//second*
     *priority attrib.*

```
   Case n:
   Priority_Attribute='Attributevalue n '// place it n
   Check InformationGain //calculate Information Gain
   Allocate the n^th Highest InformationGain // n^th priority
attrib.
   Default:
   Priority _Attribute = 'Invalid'
   }
   }
   Print("Balanced Reconstructed Tree")
   Check for path preservations;
   Generate Classification Rules;
}
```

Figure 4 (b). Prioritized Height balancing for Decision tree

Most of operations on decision trees are time consuming, which depends on height of the tree, so it is desirable to keep the height as small as suggested in C4.5 leading to less classification rules, where few attributes are missed out when the decision tree constructed. Some of the classification rules found to be undesirable for the users, resulting in scalability and efficiency problem, which will be discussed in Section 4 with an example.

## 4. PRIORITY BASED DECISION TREE CONSTRUCTION

Consider education dataset (edu_dataset) shown in Table 3, with following attributes avg_edu_level, country, and income_level with attribute values of type *{Illiterate, fouryrcollege, twoyrcollege, graduateschool, elementaryschool, Phd}, {Cuba, USA, China, India} and {HIGH, MEDIUM, LOW}* respectively.

Table 3. Attributes with distinct attribute values.

| Average Education level | Country | Income level |
|---|---|---|
| Illiterate | Cuba | Low |
| Fouryearscollege | USA | Medium |
| Fouryearscollege | USA | High |
| Fouryearscollege | USA | High |
| Twoyearscollege | USA | High |
| Graduate school | China | High |
| Elementary school | Cuba | Low |
| High school | India | Low |
| Fouryearscollege | China | Medium |
| Graduate school | China | High |
| Junior High | China | Low |
| Twoyearscollege | India | Medium |
| Fouryearscollege | USA | Medium |
| Graduate school | China | High |
| Ph. D | India | High |
| Illiterate | Cuba | Low |
| Twoyearscollege | India | Medium |

When the decision tree is constructed using C4.5/J48 classifier [12] for the edu_dataset it has pruned the attribute values of country attribute as shown in Figure 5. The Constructed decision tree has attribute values of income_level only which reduces the scalability for large datasets.

The other important aspect is efficiency is not achieved, for proper evaluation of classification rules; the resultant outputs obtained using J48 classifier of WEKA TOOL [12], reveals that priority to be allocated for attribute values at each level of abstraction when the decision tree is built.

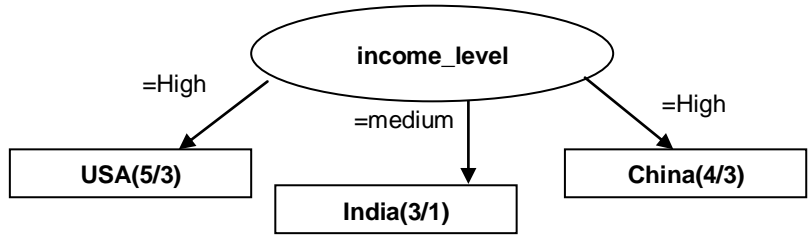

Figure 5. Decision tree using C 4.5/ J48 classifier

This is achieved with our proposed Algorithms discussed, and for simplicity the working of our algorithms shown, using modified DMQL *query* [9], with the example 4.1. The output obtained is shown in Figure 6.

**Example 4.1:** *Classification task.*

**Classify Decision_Tree**
**according to priority**{*country*(India,USA,China,Cuba) **attribute values**}
**in relevance to** *income_level*
**where attribute values for** *income_level* **count**   //leaf node(s)
**from edu_**dataset

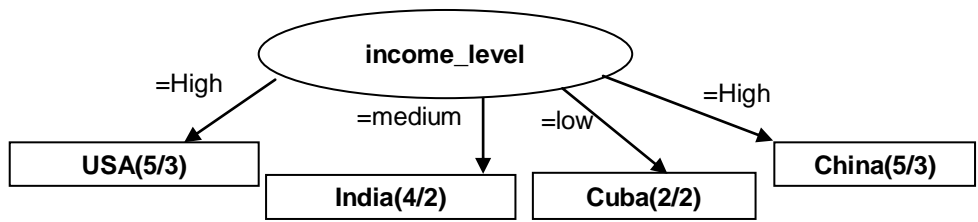

Figure 6. Decision tree with all attribute values obtained after DMQL query applied to *income_level* attribute

It is observed, some of the nodes are pruned in C4.5 classifier when decision tree, is built as shown in Figure 5. The outputs obtained after applying modified DMQL query that constitutes our proposed algorithms, for decision tree construction for the same edu_dataset, which is *far better than earlier* as shown in Figure 6, that considers all the attributes of users interest.

## 5. PERFORMANCE EVALUATION

The classifiers like C4.5, ID3 and other classifiers (SLIQ, SPRINT) use decision tree induction method which follows greedy top-down approach recursively. The decision tree induction employs Attribute selection measure like Information Gain equations (1) and (2) discussed in Section 2, that selects the test attribute at each node in the tree, during the pre-pruning and post-pruning of branches in the decision tree construction. [1], [3], [4]. Outputs obtained using WEKA tool, C4.5 classifier is shown in Figure 7.

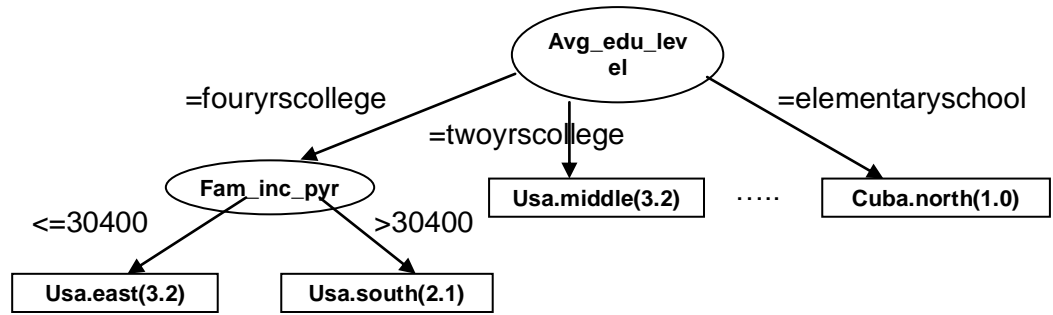

Figure 7. Decision tree with multiple (IF…THEN) rules when the attribute *avg_edu_level* is selected for education type

Attribute is represented by ellipse whereas actual attribute values are represented in rectangular box with number specified throughout the figures drawn in this paper. The Decision tree constructed in Figure 7, depicts that out of four distinct attributes i.e., *avg_edu_level* and *family_income_peryear* attributes are only shown, remaining attributes like *region* and *income_level* are pruned from the tree based on attribute selection measure of information gain. It is also conveyed that the root node selected by C4.5 classifier, is *avg_edu_level* attribute always. *Family_income_per_year* attribute can be calculated by counting the number of leaf nodes shown in rectangular boxes, with two conditions only. The remaining conditions are skipped. Overall the number of classification rules is reduced to nine in the decision tree.

Classification rules are retrieved by traversing the tree from root node to leaf nodes. To achieve this task of scalability and efficiency the *NodeMerge* and *HeightBalance* algorithms, applied by considering the concepts of priority based decision tree construction, followed by AOI (Attributed oriented induction), relevance analysis and multilevel mining technique as discussed in Section 3, at each level considering all nodes. This is achieved with modified DMQL *query* (constitutes our proposed algorithms), shown in example 5.1. The outputs obtained are shown in Figure 8.

**Example 5.1:** *Classification task.*

**Classify Decision_Tree**
**according to priority1**{*country*(India,USA,China,Cuba)
**attribute values**}
**according to priortiy2** {*region*("India.east", "India.west",
"India.south","USA.east", "USA.west", "USA.south",
"USA.middle", "Cuba.north", "Cuba.south",
"China.south","China.east","China.west") **attribute values**}
**in relevance to** *fam_inc_peryear*
**with attribute values for** *region* **count**   //leaf node(s)
**from** edu_dataset
All the Classification rules are retrieved by traversing the tree from root node to leaf nodes accurately without missing a single rule, as seen in the earlier case shown in Figure 7. (previous decision tree some of the attributes are skipped), In this way scalability for large databases with multiple attributes at each level can be achieved efficiently with less classifier errors. The Minimum Description Length (MDL) principle used to evaluate the cost of tree is discussed [13].

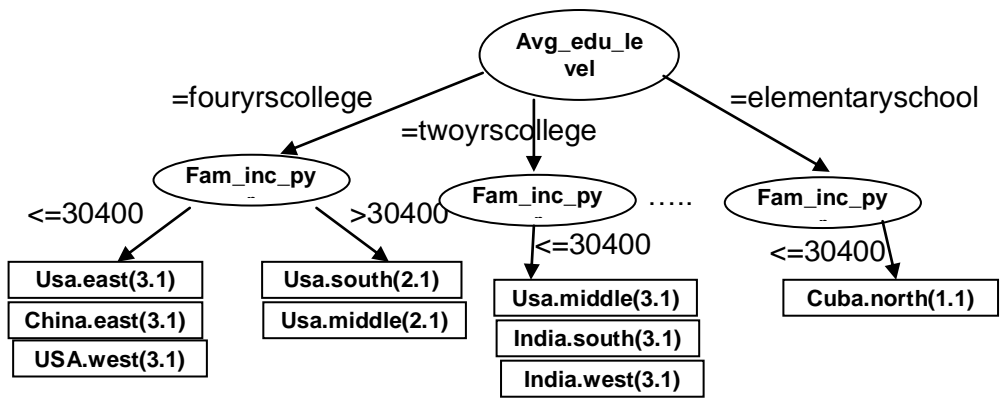

Figure 7. Priority based Decision tree based on above DMQL query

It is observed that in Figure 8, decision tree built, has considered multiple conditions for distinct attributes and attributes values given, which is of users interest. The parameters that can be compared with C4.5 classifier is shown in Table 4 with the proposed approach.

## 6. CONCLUSION

The methodology to improve C4.5 Classifier discussed, and the deficiencies exits in C4.5 classifier analyzed using WEKA tool by constructing decision trees with different types of data sets, especially with large data sets, (*for practical understanding,* in our case we have taken **T4I5K3N8** education data set) resulted in few classification rules as shown in Section 4, using C4.5 classifier.

To achieve scalability and efficiency in decision trees, many Classification rules are retrieved. *NodeMerge* and *HeightBalance* algorithms, applied by considering the concepts of priority based decision tree construction *i.e HeightBalancePriority* algorithm, as discussed in Section 3, at each level that consider all nodes. Modified DMQL *query*, is used for better understanding as discussed in Section 4 and 5 with many examples by taking education data set (edu_data set).

Performance is evaluated based on parameters mentioned in Table 4. The other parameters like classification of correctly classified instances and incorrectly classified instances can also be evaluated using precision and recall measures.

Table 4. Comparison of C 4.5 /J48 classifier with the proposed approach (*HeightBalancePriority* Algorithm).

| Parameter / Classifier | Measure used for pruning and decision tree construction | Number of attributes used for decision tree construction with priority | Multiple conditions | Selection of attribute with attribute values | Pruning and DMQL support |
|---|---|---|---|---|---|
| C4.5/J48 | Information Entropy | Few nodes without priority | Few Classification rules | Not supported | Top-down approach recursively, Pre and post pruning, no DMQL query support |
| Proposed approach with Priority | Highest Info. gain & Uncertainty co-efficient, with threshold coefficient and priority | All nodes considered with priority using *HeightBalancePriority algorithm* | Many Classification rules, helps to satisfy user specified QUERY | supported | *NodeMerge* and *Height balance* algorithms with Dynamic pruning, matching user specified DMQL query |

Our future approach is to use the constructed priority based Decision tree, for Predicting the missing and unknown values in very large data sets, and evaluate them using Confusion matrix.

**Authors**

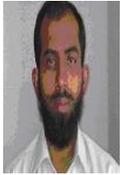 Mohd. Mahmood Ali, working as an Assistant Professor in the Dept. of C.S.E., Muffakhamjah College of Engg. & Tech., Hyderabad, India is a research scholar pursuing part-time Ph.D. from Osmania University, Hyderabad, A.P. India under the guidance of Dr. Lakshmi Rajamani, Prof CSE, College of Engg., Osmania University, Hyderabad India.
mahmoodedu@gmail.com

Mohd. S. Qaseem, working as Associate Professor in the Dept. of C.S.E / IT, Nizam Institute of Engg. & Tech., Hyderabad, India is a research scholar pursuing part-time Ph.D. from Acharya Nagarjuna University, Guntur, A.P. India under the guidance of Dr. A. Govardhan, Prof CSE , DE JNTU Hyderabad India.
ms_qaseem@yahoo.com

Dr. Lakshmi Rajamani, Professor, Dept. of C.S.E., Univ. College of Engg., O.U., Hyderabad, India. She has published more than 40 papers in reputed journals and international conferences includes IEEE, ACM, Springer, Inder science and others. She has supervised more than 7 Ph.D Scholars. She is also an active member of IEEE and ACM.
drlakshmiraja@gmail.com

Dr. A. Govardhan, Professor and Director of Evaluation, JNTU Hyderabad, India.

dejntuh@jntuh.ac.in